\documentclass[final,5p,times,twocolumn,numbers]{elsarticle}

\usepackage{amssymb}
\usepackage{amsmath}
\usepackage{booktabs}
\usepackage[table]{xcolor} 
\usepackage{hyperref}

\journal{ArXiV}

\begin{document}

\begin{frontmatter}



\title{Referring Camouflaged Object Detection With Multi-Context Overlapped Windows Cross-Attention\tnoteref{t1}} 

%

\author[1]{Yu Wen\fnref{fn1}}
\ead{20221513031@stu.sspu.edu.cn}
\author[2]{Shuyong Gao\fnref{fn1}}
\ead{sygao18@fudan.edu.cn}
\author[3]{Shuping Zhang\corref{cor1}}
\ead{spzhang@sspu.edu.cn}
\author[3]{Miao Huang\corref{cor1}}
\ead{huangmiao@sspu.edu.cn}
\author[3]{Lili Tao}
\ead{lltao@sspu.edu.cn}
\author[4]{Han Yang}
\ead{han.yang6@geely.com}
\author[5]{Haozhe Xing}
\ead{hzxing21@fudan.edu.cn}
\author[1]{Lihe Zhang}
\ead{20241513018@sspu.edu.cn}
\author[6]{Boxue Hou}
\ead{amoshou07@163.com}

\affiliation[1]{organization={School of Computer and Information Engineering, Shanghai Polytechnic University,  Shanghai 201209, China}}
\affiliation[2]{organization={Shanghai Key Laboratory of Intelligent Information Processing, School of Computer Science, Fudan University, Shanghai 200433, China}}
\affiliation[3]{organization={Department of Automation, School of Intelligent Manufacturing and Control Engineering, Shanghai Polytechnic University, Shanghai 201209, China}}
\affiliation[4]{organization={Zeekr, Geely, Shanghai 200002, China}}
\affiliation[5]{organization={Academy for Engineering and Technology, Fudan University, Shanghai 200433, China}}
\affiliation[6]{organization={Research and Design Center, Shanghai Insititue Of Computer Technology Company, Shanghai 200040, China}}

\fntext[fn1]{Co-first author}
\cortext[cor1]{Co-corresponding author}
\tnotetext[t1]{This work is supported by National Nature Science Foundation of China (Grant No. 62203291).}

\begin{abstract}
Referring camouflaged object detection (Ref-COD) aims to identify hidden objects by incorporating reference information such as images and text descriptions. Previous research has transformed reference images with salient objects into one-dimensional prompts, yielding significant results. We explore ways to enhance performance through multi-context fusion of rich salient image features and camouflaged object features. Therefore, we propose RFMNet, which utilizes features from multiple encoding stages of the reference salient images and performs interactive fusion with the camouflage features at the corresponding encoding stages. Given that the features in salient object images contain abundant object-related detail information, performing feature fusion within local areas is more beneficial for detecting camouflaged objects. Therefore, we propose an Overlapped Windows Cross-attention mechanism to enable the model to focus more attention on the local information matching based on reference features. Besides, we propose the Referring Feature Aggregation (RFA) module to decode and segment the camouflaged objects progressively. Extensive experiments on the Ref-COD benchmark demonstrate that our method achieves state-of-the-art performance.
\end{abstract}



\begin{keyword}
Binary Segmentation; Camouflaged object detection(COD); Referring camouflaged object detection(Ref-COD). 

\end{keyword}

\end{frontmatter}



\section{Introduction}
\label{sec1}
{C}{amouflaged} object detection (COD) aims to find and identify objects that are similar to their surroundings. 
This type of research can benefit the application of numerous industries, such as medical image polyp segmentation\cite{ref1}, agricultural pest detection\cite{ref2}, and industrial defect detection \cite{ref3}, etc. 
Despite the impressive achievements in this research, most studies focus on camouflaged objects within single images. While multiple information fusion methods for COD still have significant room for improvement.
This information included depth, frequency domain, reference image, and text descriptions. Methods that integrate reference prompts, such as reference images and text descriptions, are also referred to as referring camouflaged object detection (Ref-COD). This type of research not only further advances studies in COD but also promotes the advancement of multimodal human-machine interaction in more complex scenarios \cite{ref4,ref5}.
In previous work, they transformed the reference information with rich features into a one-dimensional prompt through a complex process, as shown in Figure 1(a). They utilize these low-dimensional prompts to guide the main trunk of the model in detecting camouflaged objects. Although the low-dimensional reference prompt method improves model performance, it may not fully exploit all the representative features of the reference images and often requires more images to achieve optimal performance. This situation may not be friendly in human-machine interaction.

\begin{figure}[t]
  \centering
   \includegraphics[width=1.0\linewidth]{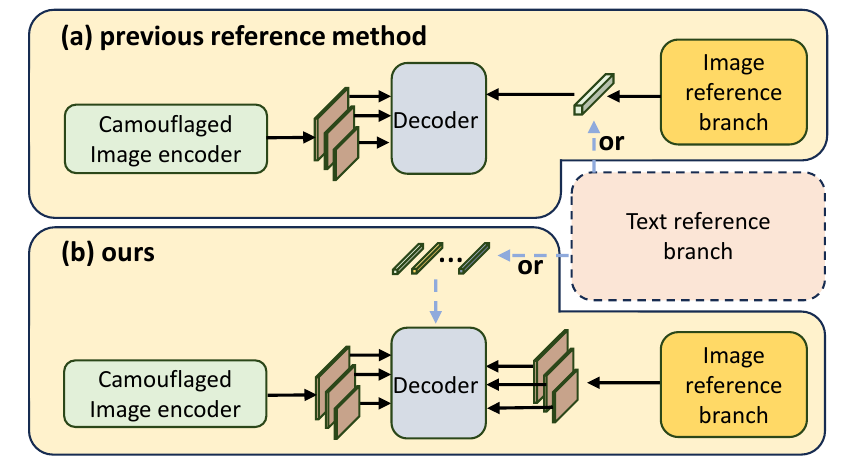}
   \caption{Comparison of previous work with our method. (a) Fuse the low-dimensional feature from the reference branch with multi-layer feature maps encoded from the camouflaged image. (b) We integrate the multi-context information from both reference features and camouflage map features.}
   \label{fig:onecol}
\end{figure}

We investigate whether we can fully leverage the features of each reference image to enable the model to achieve optimal performance with fewer reference images. Based on this idea, we propose a novel, simple, and effective network: RFMNet.
As shown in Figure 1(b), we utilize multi-context referring features to interact with the camouflaged feature maps in the decoding stage and accomplish the Ref-COD task.

The RFMNet is trained on the R2C7K dataset \cite{ref4}, which comprises camouflaged object images and their corresponding category reference salient object images. The network contains the side branch and the main branch. The side branch mainly focuses on extracting the reference features. To make the main branch fuse the reference features more flexibly and effectively, we utilize only the backbone network to encode and extract the rich, multi-contextual reference image features, thereby eliminating the need for a complicated post-processing step. 
The main branch consists of three stages: feature extraction, reference feature integration, and the decoding stage. 
We utilize the backbone network to extract camouflaged image features for the feature extraction stage. 
The reference feature integration stage mainly focuses on integrating the reference and camouflaged features. 
Drawing inspiration from the rich contour and texture details typically present in salient objects, as well as the inherent subtlety of camouflaged objects that renders them difficult to perceive, we hypothesize that effective local information matching is crucial for referring camouflaged object detection. Motivated by this notion, we propose the Overlapped Windows Cross-attention mechanism, which enables the block to concentrate on local features matching and fold the fused features back to their original size.
In the decoding stage, we propose the Referring Feature Aggregation (RFA) module, which aggregates the fused features from high to low levels in a step-by-step manner and generates the segmentation results.

Extensive experiment results demonstrate that our proposed modules can effectively improve the model performance. Furthermore, compared with the other models, our RFMNet achieves state-of-the-art performance.

In summary, our main contributions can be summarized as follows:
 
\begin{itemize}
  \item In Ref-COD tasks, we fully exploit reference image features and propose a novel network: RFMNet, which integrates multi-context reference features into camouflaged feature maps and achieves optimal performance with fewer images.
  \item We propose an overlapped windows cross-attention mechanism that enables the module to focus more on regional features matching based on reference image features, thereby effectively improving the performance of the Ref-COD model.
  \item We propose a referring feature aggregation (RFA) module, which progressively aggregates features layer by layer and generates the detailed segmentation results successfully. 

\end{itemize}

\section{Related Work}

\subsection{Camouflaged object detection (COD)}
With the development of deep learning, COD research has made rapid advancement, which can be summarized in the following categories. The mimicking animal visual mechanisms included positioning then recognition strategies \cite{ref8,ref9}, positioning then focus \cite{ref10}, zoom-in and zoom-out \cite{ref11}, and three stages localization zoom-in then recognition strategies\cite{ref12}, etc. The multi-task collaboration mechanism, such as localization, ordering, and segmentation \cite{ref13}, texture detection and segmentation \cite{ref14,ref15}, and the addition of edge-assisted detection methods \cite{ref16,ref17,ref18}. Multiple information fusion mechanisms, such as incorporating depth information \cite{ref23}, frequency-domain data \cite{ref24}, linguistic content \cite{ref4,ref5}, and images related to camouflaged object categories \cite{ref4}. This paper primarily utilizes the reference image to improve the model's performance. Through multi-context fusion and progressive decoding, our model generates satisfactory results.

\subsection{Referring Image Segmentation (RIS)}
RIS aims to segment objects based on comprehending the given reference information, predominantly in the form of text descriptions or images related to object categories.
The reference images method is also commonly known as few-shot semantic segmentation. The images requiring segmentation are designated as the query set, while the reference images are typically referred to as the support set. The majority of studies employ distinct branches to acquire features from both the support and query set, respectively \cite{ref39,ref36}. Alternatively, some studies use a shared-weight backbone to extract features \cite{ref40,ref42,ref43} and then fuse the two types of features. Due to the contradiction between SOD and COD tasks, we utilize two branches to acquire the corresponding features, respectively.

For the reference text method, a significant number of approaches primarily use a visual encoder and a text encoder to extract visual and linguistic features, respectively. Subsequently, these features are integrated to create cross-modal features, which are then fed into a decoder to generate the ultimate segmentation results. For text feature extraction, they primarily use Recurrent Neural Networks (RNNs)\cite{ref26,ref27,ref28,ref29,ref30,ref31}
or Text Transformer-based models such as BERT\cite{ref32}
and CLIP \cite{ref33,ref34} to encode the linguistic input. The fusion and decoding methods they employ are relatively flexible and diverse. These methods include concatenation \cite{ref31,ref38}, attention mechanisms \cite{ref28,ref34,ref35,ref37}, matrix multiplication \cite{ref26,ref36}, etc. Subsequently, the aforementioned methods apply convolutional layers \cite{ref31,ref35} or Transformer decoding blocks \cite{ref34,ref36,ref37,ref38} to generate the segmentation results.

\begin{figure*}[t]
  \centering
  \includegraphics[width=1.0\linewidth]{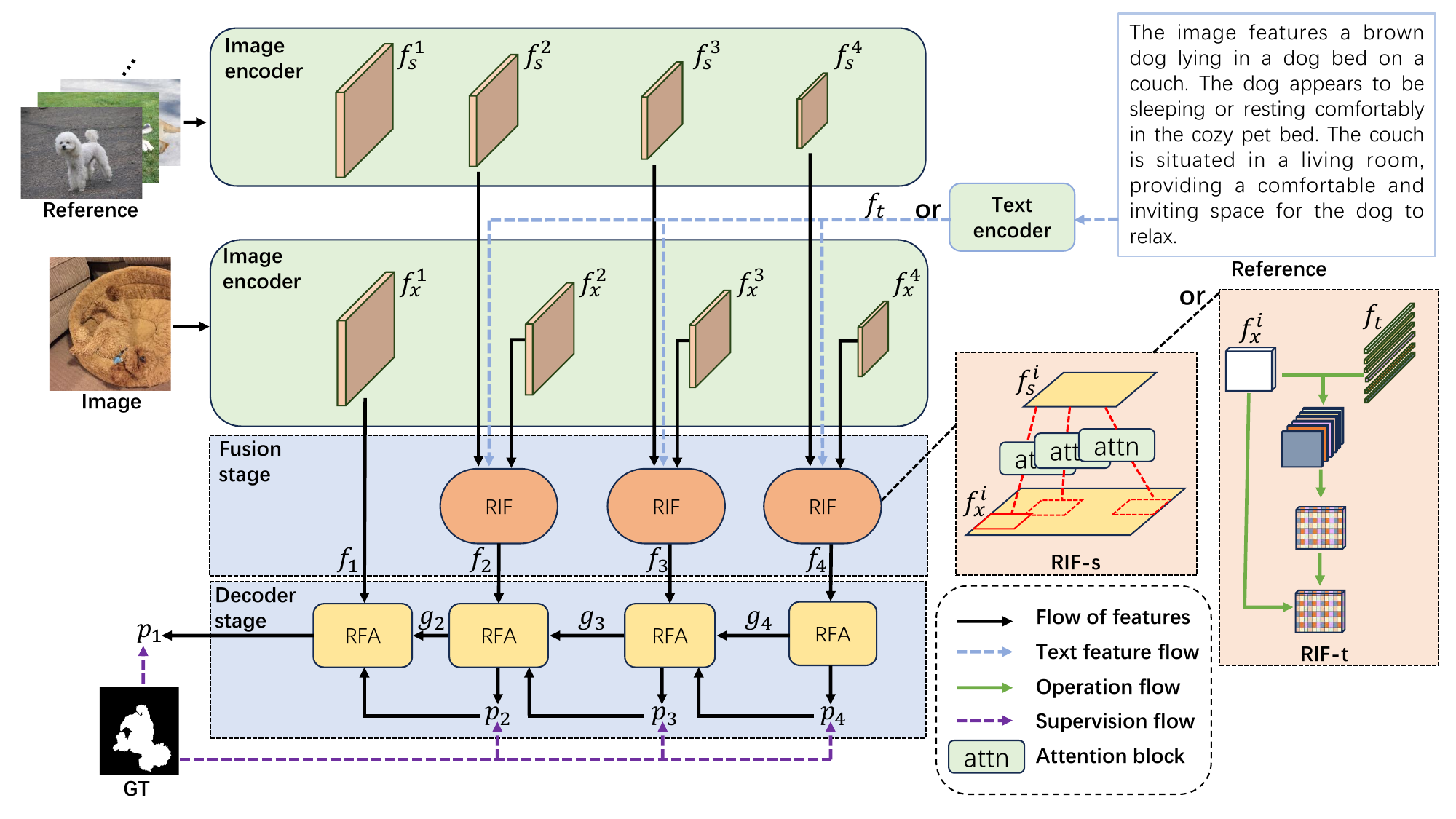}
  \caption{The overall architecture of our RFMNet. It is best viewed in color. In the feature extraction stage in green, we use the encoder to extract the camouflaged image features and the reference features, and then in the fusion stage, we use referring information fusion (RIF) modules to integrate the camouflaged features and reference features in multi-context alignment. We propose the overlapped windows cross-attention mechanism for the reference image fusion method (RIF-s). For the reference text fusion method (RIF-t), we propose a text semantics-guided referring object enhancement module. After the fusion stage, the fused features are fed into the referring feature aggregation (RFA) modules to generate the segmentation results progressively.}
  \label{fig:short}
\end{figure*}

\subsection{Referring Camouflaged object detection (Ref-COD)}
Ref-COD refers to segmenting the camouflaged objects and enhancing the detection performance based on given reference information, such as text, sound, and reference images. \cite{ref4} proposed the Ref-COD tasks, and created the R2C7K dataset along with proposing R2CNet. They proposed a referring mask generation module to fuse the two kinds of features, and subsequently, they used the referring feature enrichment module to generate the segmentation results. \cite{ref5} accomplished the task of Ref-COD in referring text based on large models. They process and encode the multi-level knowledge descriptions of the camouflaged object and scene understanding. Then, they fuse it into the visual decoding module to generate the segmentation result. In this paper, we align multi-context features and introduce an overlapped windows cross-attention mechanism to enhance local feature representation. Finally, we systematically aggregate and decode the multi-level referring features layer by layer to effectively accomplish the Ref-COD task.

\section{Proposed Method}
\subsection{Overall architecture}
The overall architecture of our RFMNet is illustrated in Figure 2. RFMNet consists of two branches for feature extraction, which extract the camouflaged image features and the reference features, respectively. The two branch features are then fed into the RIF modules in the fusion stage. For the fusion method of reference images, we propose the overlapped windows cross-attention mechanism to enable the module to pay more attention to local feature matching. We introduce a text semantics-guided referring object enhancement module for the fusion method of reference text. Finally, the fused features are fed into RIF modules to generate the segmentation result layer by layer.

\subsection{Feature extraction}
Given camouflaged objects image $ I_{\text {camo}} \in R^{C \times H \times W} $ and $ K $  pieces of reference images $ I_{ref} =\{ I^{j}_{ref} \}^{K}_{j=1}  $, $ I^{j}_{ref} \in R^{C\times H\times W} $ , where $ c=3 $ is the channels of the image, and $ H $, $ W $ are the height and width of the image.  Each of them is individually fed into the backbone network, and subsequently get the camouflaged objects image features $ F_{x} =\{ f^{i}_{x} \}^{4}_{i=1} $ and the reference features $ F^{'}_{s} =\{ F^{'j}_{s}\}^{K}_{j=1}$, $ F^{'j}_{s}  =\{ f^{ji}_{s}\}^{4}_{i=1} $ in the multi-stage process. The multi-level reference features are then concatenated and convolved to obtain the overall multi-level reference features $ F_{s}  =\{ f^{i}_{s}\}^{4}_{i=2}  $, 
\begin{equation}
  f^{i}_{s}  =Conv1(cat(\{ f^{ji}_{s}\}^{K}_{j=1}  )) 
  \label{eq:important}
\end{equation}
where cat is the concatenation function, and Conv1 represents the 1×1 convolution block, which consists of the 1×1 convolution layer, batch normalization, and ReLU activation function.
For the long reference text, we first serialize each sentence individually to obtain the text sequences $ T\in R^{N\times C^{t} } $, where N represents the number of text sequences, and $ c^{t} $ is the sequence length. After the serialization operation, we use the CLIP text encoder to get the text reference features $ F_{t} =f_{t}=\{f^{j}_{t}\}^{N}_{j=1} $.
Afterward, we feed the camouflage and reference features into the RIF module to acquire fused features.

\subsection{Referring information fusion}
In the fusion stage, the RIF module fuses camouflage and reference features.
We propose the overlapped windows cross-attention mechanism for the reference image fusion method to make the module focus on the matching local features. And given the camouflage features $ \{ f^{i}_{x}\}^{4}_{i=1}  $ and the multi-level reference image features $ \{ f^{i}_{s}\}^{4}_{i=2}  $ , the fusion process can be briefly described as follows:
\begin{equation}
f_{i} =\begin{cases}
RIF_{s}(f^{i}_{x},f^{i}_{s}),i=2,3,4. \\
f^{i}_{x},i=1. 
\end{cases}
\end{equation}

 Given the referring text features $ f_{t} $ for the reference text fusion method, we propose the text semantics-guided referring object enhancement module. The fusion process can be briefly described as follows:
\begin{equation}
f_{i} =\begin{cases}
RIF_{t}(f^{i}_{x},f_{t}),i=2,3,4. \\
f^{i}_{x},i=1. 
\end{cases}
\end{equation}
Following previous works\cite{ref9,ref55_DGNet, ref4} and considering the computational cost, we did not conduct reference features fusion in the low-level feature map $ f^{1}_{x} $.

\begin{figure}[t]
  \centering
   \includegraphics[width=1.0\linewidth]{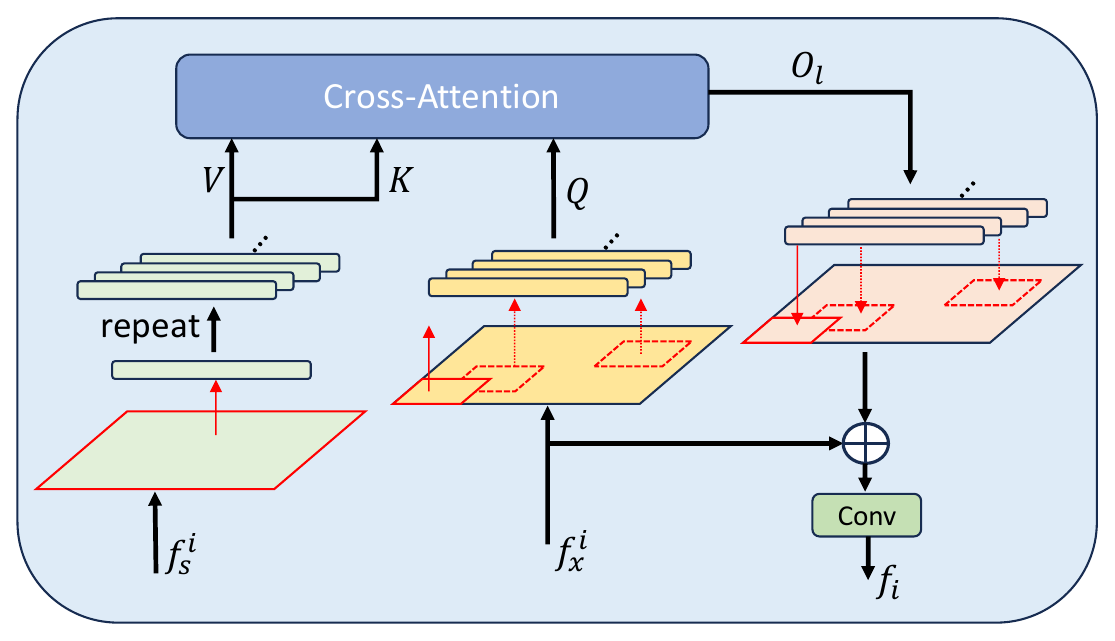}
   \caption{The reference image fusion method: overlapped windows cross-attention mechanism. Note that $ \bigoplus $ is the pixel-wise additional operation.}
   \label{fig:figure3}
\end{figure}

\subsubsection{Overlapped Windows Cross-attention mechanism}
For the reference image fusion method $ RIF_{s} $, inspired by \cite{ref44}, as well as the notable disparity in object saliency between the saliency map and the camouflage map. We propose the overlapped windows cross-attention mechanism to make the module pay more attention to the local feature matching. As shown in Figure 3, given the camouflage features $ 
\{ f^{i}_{x}\in R^{C_{i}\times H_{i}\times W_{i}    }\}^{4}_{i=2}   $ and the reference image features $ 
\{ f^{i}_{s}\in R^{C_{i}\times H_{i}\times W_{i}    }\}^{4}_{i=2}   $, where $ C_{i} $, $ H_{i} $ and $ W_{i} $ represents the channels, height and width of the feature map respectively. We divide the camouflage feature map $ f^{i}_{x} $ with overlapped windows. We set the windows size to $ k $ , the windows sliding step is $ \dfrac{k}{2} $ then the number of windows is
\begin{equation}
m=\dfrac{H_{i}-k }{k/2 }+1,
\end{equation}
and the divided windows are $ \{f^{ij}_{win}\}^{m}_{j=1}  $. Afterward, we perform the linear transformation and multi-head attention divisions on the camouflage features windows $ f^{ij}_{win} $ and reference image features $ f^{i}_{s} $.

\begin{figure}
    \centering
    \includegraphics[width=1.0\linewidth]{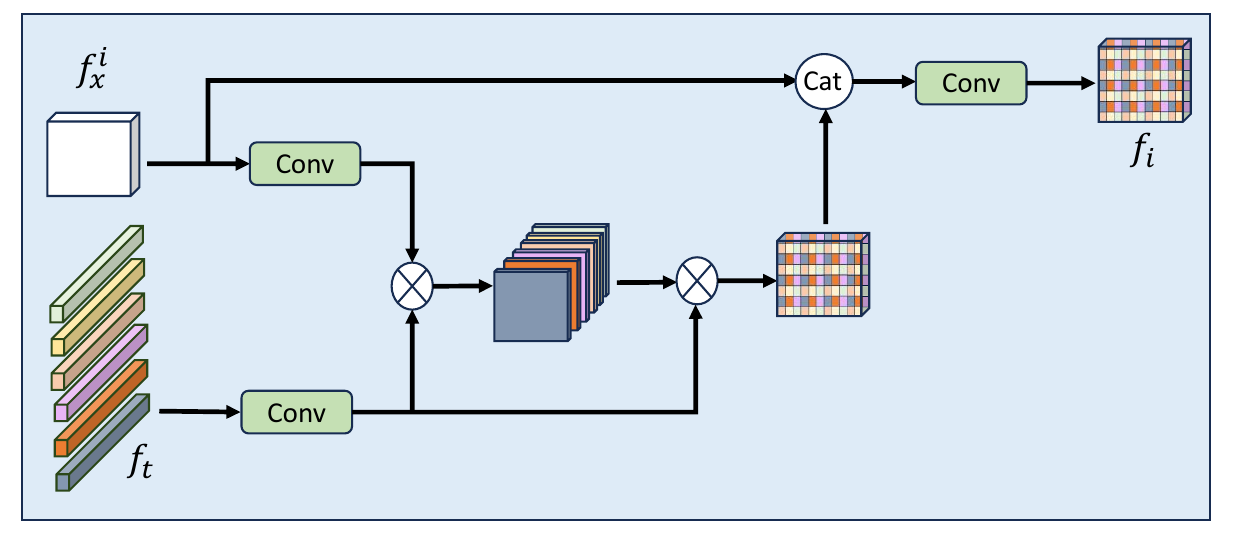}
    \caption{The text semantics-guided referring objects enhancement module. Note that ’$ Conv $’ represents the $ 1 \times 1 $ convolution block, ‘$ cat $’ is concatenation operation, 
    $ \bigotimes $ is the matrix multiplication.}
    \label{fig:figure4}
\end{figure}

\begin{equation}
q=Transpose(M H(Linear(f^{ij}_{win}))),      
\end{equation}
\begin{equation}
q\in R^{Heads\times k^{2}\times\frac{C}{Heads}  }, 
\end{equation}
\begin{equation}
k=Transpose(M H(Linear(f^{i}_{s}))), 
\end{equation}
\begin{equation}
k\in R^{Heads\times H^{2}_{i}\times \frac{C}{Heads} }, 
\end{equation}
\begin{equation}
v=Transpose(M H(Linear(f^{i}_{s}))), 
\end{equation}
\begin{equation}
v\in R^{Heads\times H^{2}_{i}\times \frac{C}{Heads} }, 
\end{equation}
where Linear represents linear transformation, $MH$ represents the multi-head decomposition. Transpose means dimension transposition. Subsequently, we compute the similarity between camouflage and reference features through cross-attention in each head.
\begin{equation}
O_{l}=Attention(q,k,v)=Softmax(\dfrac{q
k^{T} }{\sqrt{d} } )v 
\end{equation}
where $d=\dfrac{C}{Heads}$. Afterward, we transpose and perform a linear transformation on $ O_{l} $ to restore it to the shape of its original windows and fold the windows to obtain the fused interactive features. In the overlapped area of two windows, we perform the average operation to get the resulting value.
\begin{equation}
E_{i}=fold(\{O_{l}\}^{m}_{l=1}).
\end{equation}
Finally, in order to preserve the original camouflage features, residual connections and convolutions are adopted to enhance features.
\begin{equation}
f_{i}=Conv1(\alpha \cdot E_{i}+(1-\alpha)\cdot f^{i}_{x})
\end{equation}
Where $ \alpha $ is the learnable parameters, in contrast to the perspective presented in paper \cite{ref44}, where overlapped windows attention feature interaction is conducted between high-level and low-level features, our approach focuses on the interaction between reference features and the camouflage features.

\subsubsection{Text Semantics-Guided Referring Object Enhancement module}

In the reference text fusion method $ RIF_{t} $, we propose a text semantics-guided referring object enhancement module to enhance the features of the camouflage object area. Referenced \cite{ref45}, which proposed a query generation module to generate a set of query vectors by emphasizing different words. Our module generates the object enhancement vectors by emphasizing the importance of different sentence descriptions. As shown in Figure 4, we conduct the matrix multiplication with the camouflage features $ f^{i}_{x} $ and the reference text features $ f_{t} $ to generate the attention weights for different sentences. Next, the sentence attention weights are applied to the text semantic features to get the camouflaged object enhancement vectors. In order to maintain the original camouflage features, the concatenation and convolution block are applied to get the final referring camouflaged object feature maps $ f_{i} $.

\begin{table*}[t]
  \centering
  \caption{Comparison of the COD methods and its relative Ref-COD models, 'Overall': all the camouflaged objects images, 'Single-object': images with single camouflaged object, 'Multi-objects': images with multiple camouflaged objects, '-S': COD model with salient object images reference, '-T': COD model with text reference, 'N': number of reference images, 'R-50': Resnet-50, 'R2-50': Res2Net-50, 'E-B4': EfficientNet-B4, 'SF-B4': SegFormer-B4, 'PVTv2': PVTv2\cite{ref58}, 'Swin-S': SwinTransformer-S, '$\uparrow$': the higher the better, '$\downarrow$': the lower the better. \textbf{Bold} and \underline{underline} represent the first and second best results respectively.
  }
  \small
  \begin{tabular}{p{2.7cm}p{0.3cm}p{1.0cm}p{0.01cm}p{0.5cm}p{0.6cm}p{0.5cm}p{0.6cm}p{0.01cm}p{0.5cm}p{0.6cm}p{0.5cm}p{0.6cm}p{0.01cm}p{0.5cm}p{0.6cm}p{0.5cm}p{0.6cm}}
    \toprule
     Models & N & Backbone & & \multicolumn{4}{c}{Overall}   &       & \multicolumn{4}{c}{Single-object} &       & \multicolumn{4}{c}{Multi-objects} \\
\cmidrule{5-8}\cmidrule{10-13}\cmidrule{15-18}          &   &   & &  $S_{\alpha}\uparrow$ & $\alpha$E$\uparrow$ &  $F^{\omega }_{\beta }\uparrow$ &  M $\downarrow$  &       &  $S_{\alpha}\uparrow$ & $\alpha$E$\uparrow$ &  $F^{\omega }_{\beta }\uparrow$ &  M $\downarrow$   & &  $S_{\alpha}\uparrow$ & $\alpha$E$\uparrow$ &  $F^{\omega }_{\beta }\uparrow$ &  M $\downarrow$\\
    \midrule
    R2CNet-T\cite{ref4}	& - &R-50 &&0.806	&0.878	&0.668	&0.037& &	0.810	&0.880	&0.674	&0.035	& &	0.753	&0.870	&0.607&	0.046 \\ 
    \rowcolor{gray!10}
    R2CNet-S\cite{ref4}	& 5 & R-50 && 0.805	& 0.879 & 0.669	& 0.036 & &		0.810 &	0.880	& 0.674	& 0.035	& &	0.747	& 0.872	& 0.602 &	0.046\\
    PFNet-T\cite{ref4, ref10} & - &R-50 &&0.813	&0.893	&0.693	&0.034	&&	0.817&	0.892	&0.697	&0.033	&&	0.769	&0.889	&0.648	&\underline{0.041} \\ 
    \rowcolor{gray!10}
    PFNet-S\cite{ref4, ref10}	& 5 &R-50 &&0.811	&0.885	&0.687	&0.036	&&	0.815	&0.886	&0.691	&0.035	&&	0.764	&0.873	&0.632	&0.045 \\
    PreyNet-T\cite{ref4, ref54_PreyNet} & - &R-50 &&0.816	&\underline{0.901}	&0.705	&0.033	&&	0.821	&\underline{0.900}	&0.710&	0.032	&&	0.759&	\textbf{0.902}&	0.648	&\underline{0.041}\\
    \rowcolor{gray!10}
    PreyNet-S\cite{ref4, ref54_PreyNet} & 5 &R-50	&&0.817	&0.900	&0.704	&0.032	&&	0.822&	\underline{0.900}	&0.709&	0.032	&&	0.763	&0.898	&0.645	&\underline{0.041}\\
    SINetV2-T\cite{ref4, ref9} & - &R2-50 &&0.822	&0.887	&0.696	&0.033	&&	0.827&	0.888	&0.702	&0.032	&&	0.766&	0.866&	0.629	&0.043\\
    \rowcolor{gray!10}
    SINetV2-S\cite{ref4, ref9}	& 5 &R2-50 &&	0.823	&0.888	&0.700&	0.033	&&	0.828	&0.889&	0.705&	0.032	&&	0.771	&0.874&	0.634&	0.043\\
    DGNet-T\cite{ref4, ref55_DGNet}	& - &E-B4 &&0.824&	0.891&	0.701	&0.032&&		0.830	&0.892	&0.709&	0.031	&&	0.745	&0.873	&0.596	&0.046\\
    \rowcolor{gray!10}
    DGNet-S\cite{ref4, ref55_DGNet} & 5 &E-B4 &&	0.821	&0.891	&0.696	&0.032	&&	0.827	&0.890	&0.703&	0.031	&&	0.748&	0.879&	0.607	&0.045\\

    RFMNet-T(ours) & - &	R-50 &&	\underline{0.827}&	0.899&	\underline{0.718}&	\underline{0.031}&& \underline{0.831} & 0.899 & \underline{0.723} & \underline{0.030} && \underline{0.776} & 0.9 & \textbf{0.67} & \underline{0.041}\\
    \rowcolor{gray!10}
    RFMNet-S(ours) & 3 &	R-50 &&\textbf{0.829}& \textbf{0.903}& \textbf{0.719} & \textbf{0.030} && \textbf{0.833} & \textbf{0.904} & \textbf{0.725} & \textbf{0.029} && \textbf{0.781} & \underline{0.901} & \underline{0.665} & \textbf{0.04}\\
    \hline
    
    \rowcolor{gray!10}
    UAT\cite{ref57} & 5 & PVTv2 && 0.855 & 0.912 & 0.757 & 0.026 && 0.859 & 0.913 & 0.761 & 0.025 && 0.805 & 0.900 & 0.701 & \textbf{0.033} \\
    \rowcolor{gray!10}
    RPMA-S\cite{ref4, ref56} & 10  & SF-B4 && 0.862 & 0.930 & 0.784 & 0.023 && 0.867 & \textbf{0.934} & 0.791 & 0.023 && 0.806 & 0.894 & 0.718 & \textbf{0.033} \\
    \rowcolor{gray!10}
    RFMNet-S(ours) & 3 & Swin-S && \textbf{0.875} & \textbf{0.933} & \textbf{0.796} & \textbf{0.021} && \textbf{0.88} & 0.933 & \textbf{0.801} & \textbf{0.02} && \textbf{0.816} & \textbf{0.931} & \textbf{0.736} & \textbf{0.033} \\
    \bottomrule
  \end{tabular}

  \label{tab: comparision }
\end{table*}

\subsection{Referring feature aggregation module}

Inspired by \cite{ref12}, we propose the referring feature aggregation (RFA) module, which aims to aggregate the referring features from adjacent layers and produce the segmentation results. As shown in Figure 5, given the referring features $ \{ f_{i}\in R^{C_{i}\times H_{i}\times W_{i}   }  \}^{4}_{i=1} $ , the aggregation process can be briefly describe as follows:
\begin{equation}
g_{i} =\begin{cases}
Conv3(Conv3(Conv3(f_{i}))),i=4 \\
Conv3(Conv3(Conv3(k_{i}))),i=1,2,3
\end{cases},
\end{equation} 
Where Conv3 is the 3×3 convolution block, consisting of a 3×3 convolution layer, batch normalization, and ReLU activation function, moreover, the intermediate features $ k_{i} $ are obtained by the aggregation of enhanced foreground features from the previous layer $ j_{i} $ and the current features $ f_{i} $. This process can be formulated as follows:
\begin{equation}
k_{i}=Conv3(Cat(Conv3(f_{i}),j_{i})),
\end{equation}
\begin{equation}
j_{i}=Conv3(\odot(BI(g_{i+1}),BI(p_{i+1}))),
\end{equation}
Where Cat is the concatenation operation, $ \odot $ is the pixel-wise multiplication, and BI is the bilinear interpolation operation. Finally, we conduct a prediction to obtain the segmentation result:
\begin{equation}
p_{i}=C1(Conv3(g_{i})),
\end{equation}
where C1 is the $ 1\times1 $ convolution layer, in which the output channels setting is 1.

\begin{figure}
    \centering
    \includegraphics[width=1.0\linewidth]{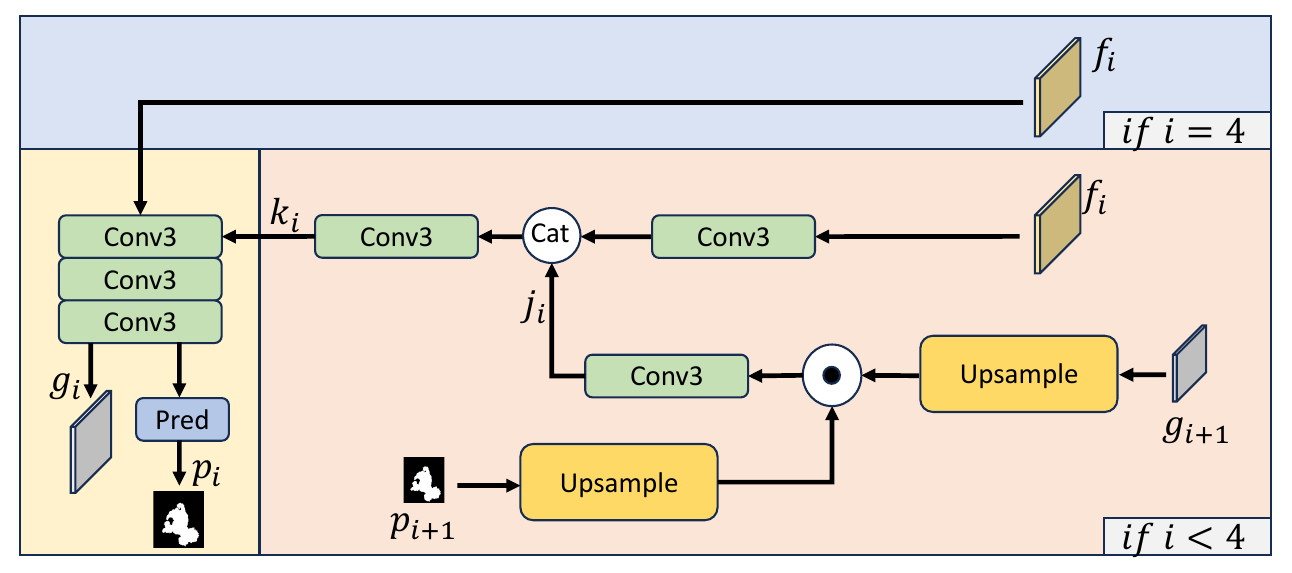}
    \caption{The referring feature aggregation module. Note that ‘$ cat $’ represents the concatenation, ‘$ Conv3 $’ is $ 3\times3 $ convolution block, $ \bigodot $ is the pixel-wise multiplication.}
    \label{fig:FIGURE5}
\end{figure}

\section{Experiments}

\subsubsection{Dataset} Our Ref-COD experiments were conducted on the R2C7K dataset \cite{ref4}, which consists of the Camo-subset and the Ref-subset. The Camo-subset includes 5,015 images of camouflaged objects from 64 different categories. Moreover, the Ref-subset is the reference image of salient objects from 64 categories, each category consists of 25 images, and the total number of Ref-subset is 1600. In order to conduct the referring text camouflaged objects detection task conveniently, we utilize the large model mPLUG-Owl2\cite{ref61} to generate text descriptions of the Camo-subset images. 

\subsubsection{Implementation details} 

We train RFMNet in two stages. In the first stage, we train the baseline model, which consists of a Backbone network and an FPN\cite{ref53_FPN} decoder, on the training set with a batch size of 32 for 45 epochs. Meanwhile, to enhance the extraction of salient object features, we utilize the DUTS\cite{ref60} dataset to train a baseline model of the same type for 45 epochs. Additionally, for the reference text method, we utilize the pre-trained CLIP to extract text features. After that, we extract the well-trained backbone network to act as the side branch. In the second stage, we train our proposed RFMNet for 500 epochs with the backbone parameters frozen. The entire model is optimized using the Adam optimizer with a polynomial decay strategy and a momentum of 0.9. The learning rate is initialized at 1.5e-4, and the power is set to 0.9. The input image is resized to 512×512. All experiments were conducted using PyTorch on a single NVIDIA GeForce RTX 4090 system.

\subsubsection{Loss function}For the prediction $ \{p_{i}\}^{4}_{i=1} $ we set up its corresponding loss $ \{\mathcal{L}_{i}\}^{4}_{i=1} $ for supervision. Considering that the weighted intersection-over-union loss $ \mathcal{L}^{\omega}_{iou}  $ and the weighted binary cross entropy loss $ \mathcal{L}^{\omega}_{bce}  $ are widely used in the COD task for the global supervision and the local regional restriction, respectively, the $ \mathcal{L}_{i} $ can be formulated as
\begin{equation}
\mathcal{L}_{i} = \mathcal{L}^{\omega}_{bce} + \mathcal{L}^{\omega}_{iou},
\end{equation}

To supervise the high-resolution predictions more effectively, we balance the weights of multiple prediction losses, and the overall prediction loss $ \mathcal{L}_{total} $ can be formulated as follows:
\begin{equation}
\mathcal{L}_{total} =7*\mathcal{L}_{1}+(4*\mathcal{L}_{2}+3*\mathcal{L}_{3}+2*\mathcal{L}_{4}).
\end{equation}
During backpropagation, $\mathcal{L}_{2}$, $\mathcal{L}_{3}$, and $\mathcal{L}_{4}$ are computed and summed first, their sum is subsequently combined with $\mathcal{L}_{1}$ to calculate the total loss $\mathcal{L}_{total}$.


\subsubsection{Metrics} Following previous evaluation metrics in COD task, we evaluate our method by four widely used metrics including Structure-measure $ (S_{\alpha}) $ \cite{ref48}, adaptive E-measure($ \alpha $E) \cite{ref49}, weighted F-measure ($ F^{\omega }_{\beta } $) \cite{ref50} and mean absolute error (M) \cite{ref51}. $ (S_{\alpha}) $ \cite{ref48} is used to evaluate the structural similarity between the prediction mask and the ground truth, $ \alpha$E \cite{ref49} focuses on the similarity evaluation in both local and global regions. $ F^{\omega }_{\beta } $ \cite{ref50} is a comprehensive assessment in both precision and recall. M \cite{ref51} is a metric measuring the absolute difference at the pixel level.

\subsection{Comparison and evaluation}

\subsubsection{Quantitative Comparison} As shown in Table 1, compared with previous Ref-COD methods, our RFMNet-S method incorporates ResNet-50 backbones achieves significant improvements. Specifically, when compared with the second-best referring salient image model DGNet-S, our method shows a 6.25\% average performance gain in terms of M and a 3.30\% improvement in $ F^{\omega }_{\beta } $. Meanwhile, compared with the referring text models, our RFMNet-T also outperforms the second-best model by 3.13\% on M. Furthermore, when compared with Ref-COD methods employing Transformer backbones, our proposed SwinTransformer-S-based RFMNet-S surpasses the second-best model RPMA-S by 1.5\% in terms of $ F^{\omega }_{\beta } $. Notably, RFMNet-S requires only three reference images, making it more user-friendly than RPMA-S. 

\subsubsection{Qualitative Evaluation} As shown in Figure 6, we present camouflage images across five challenging scenarios: small object, large object, multiple objects, occlusion, and uncertainty. In comparison to state-of-the-art models, our RFMNet-S, utilizing ResNet-50, demonstrates a remarkable capability for accurately segmenting camouflaged objects. Our model excels at precisely locating the object while effectively filtering out irrelevant areas, particularly in scenarios involving small object (e.g., the first and second columns) and uncertain situations (e.g., the ninth and tenth columns). This enhanced performance can be attributed primarily to the overlapped windows cross-attention mechanism, which emphasizes relevant features of the object while diminishing attention on uncertain regions through local semantic similarity comparisons. Furthermore, by leveraging  (RFA) modules that systematically aggregate features layer by layer, our proposed model achieves finer segmentation of camouflaged objects.

\begin{figure*}[t]
    \centering
    \includegraphics[width=1\linewidth]{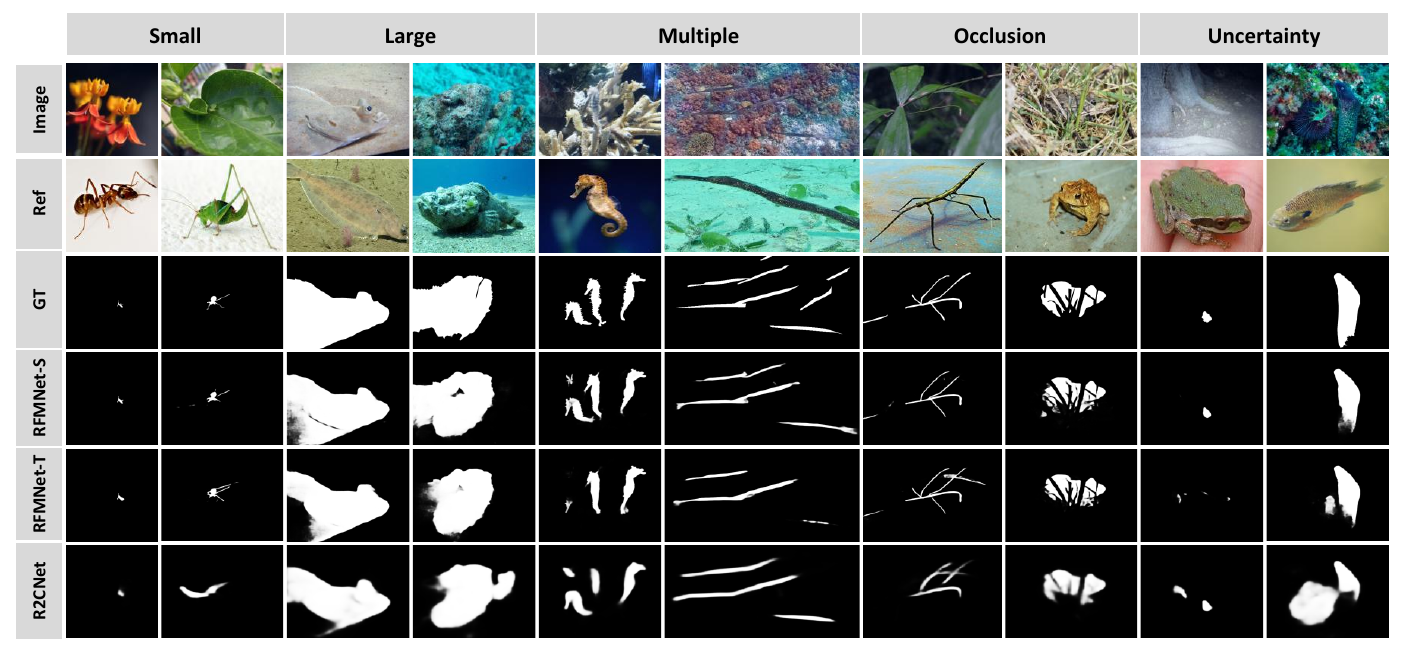}
    \caption{Visual comparison of the our proposed RFMNet with other representative Ref-COD method on five challenging scenarios. ‘RFMNet-S’: RFMNet-S with ResNet-50. 'RFMNet-T': RFMNet-T with ResNet-50. Please zoom in for more details.}
    \label{fig:figure6}
\end{figure*}

\subsection{Ablation study}
To validate the effectiveness of our proposed method, we conduct ablation experiments on the R2C7K datasets.

\begin{table}
  \centering
  \caption{Ablation experiments on the components of our proposed RFMNet. 
  }
  \small
  \begin{tabular}{p{2.2cm}p{1cm}p{1cm}p{1cm}p{0.7cm}}
    \toprule
     Components &  $S_{\alpha}\uparrow$ & $\alpha$E $\uparrow$ &  $F^{\omega }_{\beta }\uparrow$ &  M $\downarrow$ \\
    \midrule
    R50 & 0.773 &  0.835 &  0.596 &  0.045 \\
    R50+RIF$_{t}$ & 0.812 & 0.873 & 0.677 & 0.036\\
    R50+RIF$_{s}$ & 0.798 & 0.886 & 0.664 & 0.037\\
    R50+RFA & 0.822 & 0.895 & 0.705 & 0.032\\
    R50+RIF$_{t}$+RFA & 0.827 & 0.899 & 0.718 & 0.031\\ 
    R50+RIF$_{s}$+RFA & \textbf{0.829} & \textbf{0.903} & \textbf{0.719} & \textbf{0.03}\\ \hline
    SwinS+RFA & 0.87 & 0.928 & 0.785 & 0.023 \\
    SwinS+RIF$_{s}$+RFA & \textbf{0.876} & \textbf{0.933} & \textbf{0.797} & \textbf{0.021} \\
    \bottomrule
  \end{tabular}
  \label{tab: component ablation}
  

\end{table}

\subsubsection{Component Analysis} As shown in Table 2, we conduct the ablation experiments on components of RFMNet. As seen in the 2nd, 3rd, 4th rows, the referring information fusion (RIF$_{t}$ and RIF$_{s}$) module and the referring feature aggregation (RFA) module perform significantly better than the Baseline model (R50), which only has the ResNet-50 Backbone and the FPN\cite{ref53_FPN} decoder. When we combine the two modules(R50+RIF$_{t}$+RFA and R50+RIF$_{t}$+RFA), the RIF$_{t}$ and RIF$_{s}$ module also increases performance effectively. In addition, when we replace the R50 to SwinTransformer-S, the RIF$_{s}$ also improves the segmentation performance, by 0.69\%, 0.54\%, 1.53\%, 8.70\% respectively in $S_{\alpha}$, $\alpha$E, $F^{\omega }_{\beta }$, M compared with the model with SwinTransformer-S and RFA modules(e.g., 7th, 8th rows). These results indicate that the overlapped windows cross-attention mechanism and the feature aggregation module can effectively locate and segment the camouflaged object. 

\subsubsection{Number of reference images} Considering that the feature extraction method of the reference branch proposed by us differs from other existing methods, we evaluate the number of reference images in Table 3. In each training iteration, we randomly select N images as the reference images, where the transformation of N ranges from 0 to 5. These results indicate that our proposed feature extraction method and the fusion method are effective, as the best results can be achieved with only three reference images, while R2CNet\cite{ref4} and UAT\cite{ref57} employ ICON\cite{ref59} to extract reference features, requiring five images to achieve the best results. Our proposed referring feature extraction method not only proves that it can improve the model's ability to detect camouflaged objects, but also provides sufficient reference information in a considerable number of images. Although there is a commonly acknowledged agreement in theory that a greater number of reference images leads to better model performance results, for the Ref-COD task, a proper number of images also turns out to be beneficial in terms of human-computer interaction and the computational load of the model's reference feature extraction section.

\subsubsection{Partitioned windows size and step analysis}  Ablation experiments on partitioned attention window size and step size are shown in Table 4. We set the different attention window sizes and step sizes for various layers. Moreover, the experiment results indicate that the most effective way is to fuse the feature layers directly by cross-attention when the overlapped windows areas of each feature layer are common (e.g., 1st, 2nd, 3rd rows). When the size of partitioned windows differs in each layer, the most effective approach is to partition the windows of the feature layers from high to low into progressively larger regions, with the step size best set to half the size of the window (e.g., 6th row). One possible explanation is that such window size and step size settings enable the model to locate the objects more accurately based on semantic feature matching. When the fused features are aggregated with low-level features, the low-level features can perform more detailed segmentation based on attention matching of larger windows.

\begin{table}
  \centering
  \caption{Ablation experiments on the number of reference images. 
  }
  \small
  \begin{tabular}{p{1cm}p{1cm}p{1cm}p{1cm}p{0.7cm}}
    \toprule
     N &  $S_{\alpha}\uparrow$ & $\alpha$E $\uparrow$ &  $F^{\omega }_{\beta }\uparrow$ &  M $\downarrow$ \\
    \midrule
    0 & 0.822 & 0.895 & 0.705 & 0.032 \\
    1 & 0.826 & 0.898 & 0.713 & 0.031 \\
    2 & 0.828 & 0.898 & 0.716 & 0.031 \\
    3 & \textbf{0.829} & 0.903 & \textbf{0.719} & \textbf{0.03}  \\
    4 & 0.828 & \textbf{0.904} & 0.718 & 0.031 \\
    5 & 0.827 & 0.903 & 0.717 & 0.031 \\
    \bottomrule
  \end{tabular}
  \label{tab: reference images number ablation}
  

\end{table}

\begin{figure*}[t]
    \centering
    \includegraphics[width=0.95\linewidth]{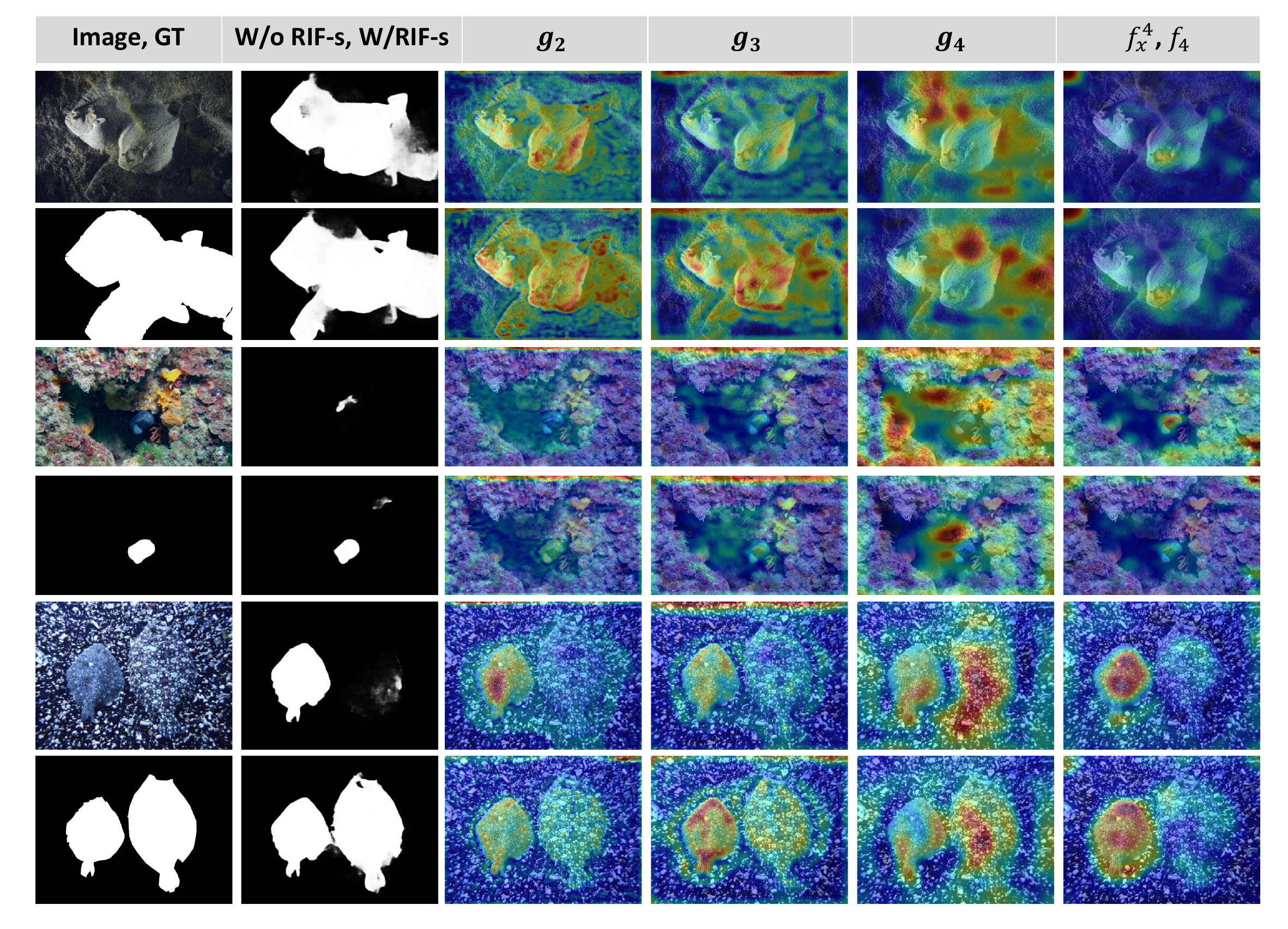}
    
    \caption{Visualization of intermediate features between RFMNet with RIF-s and without RIF. The content before ',' is the label of an odd number of rows of pictures, and the content after ',' is the label of an even number of rows of pictures.}
    \label{fig:figure7}
\end{figure*}

\subsubsection{Referring layers analysis} Table 5 compares the experimental results of adding reference information fusion to different layers. The experimental results indicate that all the referring layers can boost the performance of our model, especially the $f_{4} $ fusion layer, which improved the $F^{\omega }_{\beta }$ score from 0.705 to 0.717. When we combine the referring layers, the performance also improves. This situation is because we set different window sizes for different feature layers, and the reference information for each layer is different; the lower reference layer contains more detailed features. This multi-context fusion method thoroughly explains the advantages of different reference feature layers.

\begin{table}
  \centering
  \caption{Ablation experiments on attention window size and its step size of the RFMNet. Note that '$H_{i} $' and '$s_{i} $' are the height and the attention window size of image features '$f^{i}_{x} $' respectively. 'step' is the step size. '$s_{i}/2$' means when the attention window size is less than $H_{i} $, the step size is $s_{i}/2$. }
  \footnotesize
  \begin{tabular}{p{0.2cm}p{0.4cm}p{0.4cm}p{0.4cm}p{0.4cm}p{0.6cm}p{0.6cm}p{0.6cm}p{0.6cm}}
    \toprule
     No. & $ s_{2} $ & $ s_{3} $ & $ s_{4} $ & step & $S_{\alpha}\uparrow$ & $\alpha$E $\uparrow$ &  $F^{\omega }_{\beta }\uparrow$ &  M $\downarrow$ \\
    \midrule
    1 & $H_{2} $ & $H_{3} $ & $H_{4} $ & $s_{i} $ & 0.827 & \textbf{0.903} & \textbf{0.719} & 0.031 \\
    2 & $H_{2}/2  $  & $H_{3}/2  $ & $H_{4}/2  $  & $s_{i}/2  $  & 0.826 & 0.899 & 0.712 & 0.032 \\
    3 & $H_{2}/4  $ & $H_{3}/4  $ & $H_{4}/4  $ & $s_{i}/2  $& 0.827 & 0.897 & 0.716 & 0.031 \\
    4 & $H_{2}/4  $ & $H_{3}/2  $ & $H_{4} $ & $s_{i}/2  $& 0.825 & 0.896 & 0.709 & 0.032 \\
    5 & $H_{2} $ & $H_{3}/2  $ & $H_{4}/4  $ & $s_{i} $& 0.827 & 0.897 & 0.711 & 0.032 \\
    6 & $H_{2} $ & $H_{3}/2  $ & $H_{4}/4  $ & $s_{i}/2  $ & \textbf{0.829} & \textbf{0.903} & \textbf{0.719} & \textbf{0.03} \\
    
    \bottomrule
  \end{tabular}
  \label{tab: windows size and stride ablation}

  
\end{table}

\begin{table}
  \centering
  \caption{Ablation experiments on referring layers of the RFMNet.}
  \small
  \begin{tabular}{p{0.5cm}p{0.5cm}p{0.5cm}p{0.5cm}p{0.7cm}p{0.7cm}p{0.7cm}p{0.7cm}}
    \toprule
     No. & $ f_{2} $ & $ f_{3} $ & $ f_{4} $ &  $S_{\alpha}\uparrow$ & $\alpha$E $\uparrow$ &  $F^{\omega }_{\beta }\uparrow$ &  M $\downarrow$ \\
    \midrule
    0 & & & & 0.822 & 0.895 & 0.705 & 0.032 \\
    1 & $ \surd $ &  &  & 0.825 & 0.898 & 0.711 & 0.032 \\
    2 & & $ \surd $ &   & 0.826 & 0.897 & 0.712 & 0.032 \\
    3 & & & $ \surd $ & 0.827 & 0.902 & 0.717 & 0.031 \\
    4 & $ \surd $ & $ \surd $ &   & \textbf{0.829} & 0.9 & 0.716 & 0.031 \\
    5 & & $ \surd $ & $ \surd $ & \textbf{0.829} & 0.901 & 0.717 & 0.031 \\
    6 & $ \surd $ & & $ \surd $ & 0.827 & \textbf{0.903} & 0.717 & 0.031 \\
    7 & $ \surd $ & $ \surd $ & $ \surd $ & \textbf{0.829} & \textbf{0.903} & \textbf{0.719} & \textbf{0.03} \\
    \bottomrule
  \end{tabular}
  \label{tab: reference layer ablation}

  
\end{table}



  

\subsection{Feature visualization}

The visualization of intermediate features between models with RIF-s and without RIF modules is shown in Figure 7. Each two lines of images is an example of the detection result. The first line shows the segmentation result and intermediate feature visualization without RIF, and the second line shows the segmentation result and intermediate feature visualization with RIF-s.

\begin{table*}
  \centering
  \caption{Comparison of parameters, MACs, and Speed across Ref-COD models. All experiments are conducted on a system with an NVIDIA RTX 4090 GPU. }
  \small
  \begin{tabular}{p{0.2cm}p{2.2cm}p{1.4cm}p{0.5cm}p{0.8cm}p{0.8cm}p{0.8cm}p{0.8cm}p{1.2cm}p{1.2cm}p{1.2cm}}
    \toprule
   No. & Models & Backbone &  Size & $S_{\alpha}\uparrow$ & $\alpha$E $\uparrow$ &  $F^{\omega }_{\beta }\uparrow$ & M$\downarrow$ & Params(M)$\downarrow$ & MACs(G)$\downarrow$ & Speed(FPS)$\uparrow$   \\
    \midrule
  1 & R2CNet   & ResNet-50 & 352 & 0.805	& 0.879 & 0.669	& 0.036 & 25.1 & \textbf{11.68} & \textbf{176.73} \\
  2 & RFMNet-S(ours) & ResNet-50 & 352 & 0.801 & 0.881 & 0.664 & 0.036 & \textbf{24.72}& 13.4 & 16.7  \\
  3 & RFMNet-S(ours) & ResNet-50 & 384 & 0.807 & 0.881 & 0.676 & 0.035 & \textbf{24.72}& 15.43 & 63.28  \\
  4 & RFMNet-S(ours) & ResNet-50 & 512 & 0.829 & 0.903 & 0.719 & 0.03 & \textbf{24.72} & 27.48 & 46.79  \\
    5 & UAT & PVTv2 & 352 & 0.855 & 0.912 & 0.757 & 0.026 & 98.05 & 99.15 &  71.50 \\
  6 & RFMNet-S(ours) & Swin-S & 352 & 0.85 & 0.913 & 0.748 & 0.026 & 34.1 & 17.27 & 14.08  \\
  7 & RFMNet-S(ours) & Swin-S & 512 & \textbf{0.875} & \textbf{0.933} & \textbf{0.796} & \textbf{0.021} & 34.1 & 35.66 &  31.44 \\
    \bottomrule
  \end{tabular}
  \label{tab: reference layer ablation}
\end{table*}

{\bf{Location of camouflaged objects:}}
The overlapped windows cross-attention mechanism can filter out irrelevant areas and locate camouflaged objects. From the comparison of the results of the fourth and third lines, it can be seen that the environment around the camouflaged object will have a specific interference with the model detection. The feature extraction module pays excessive attention to the non-camouflaged object area, which renders the referring feature aggregation (RFA) module unable to properly separate the camouflaged object area from the interference region, ultimately causing the model to fail in segmenting the camouflaged object. The overlapped window cross-attention mechanism can suppress the irrelevant area based on local features matching the camouflage map, thereby reducing the attention to the area and highlighting the camouflage objects. This enables the referring feature aggregation module to gradually locate and segment camouflage objects due to the reduction of the interference region in the feature map.

{\bf{Segmentation integrity of camouflaged objects:}} As can be seen from the first row and the fifth row of Figure 7, the feature map without reference feature fusion has a large gap between the edge attention and the center attention of the high-interest area, which makes the RFA module think that ignores the potential camouflaged object area of the highly concerned edge, resulting in incomplete segmentation results. The reference and camouflage objects in the R2C7K\cite{ref4} dataset are not highly similar, and there are still differences in type details and image shading. The overlapped windows cross-attention mechanism primarily calculates the feature similarity of local regions in the camouflage image based on the reference features. After the feature is fused, the attention to the potential camouflage area is improved due to the matching of feature similarities. Compared to feature maps with uneven attention, feature maps with relatively uniform attention make it easier for the RFA module to comprehensively identify and segment camouflage regions, thereby improving the model's performance in terms of mean absolute error (M).

\subsection{Limitation}

Although our proposed RFMNet has achieved remarkable performance, it still has some inherent limitations.

As shown in Table 6. Firstly, in comparison to R2CNet, our proposed RFMNET exhibits a comparable number of parameters. However, a notable disparity exists between the two models with respect to multiply-accumulate operations (MACs) and processing speed, measured in frames per second (FPS). In particular, the computational speed of R2CNet is 3.7 times faster than that of our proposed RFMNet, as shown in the first and fourth rows. Secondly, our proposed RFMNet-S is better suited for input images of specific sizes. For an input image with a size of 352, after feature extraction, the sizes of $f^{3}_{x}$ and $f^{4}_{x}$  are {22, 11}, respectively. The pixel size and pixel step size of the divided window can only be set to 2 and 1, respectively; otherwise, edge padding is required to accommodate larger window stride settings. As a result, this input image size leads to lower computational speed. In contrast, for an input image of size 384, after feature extraction, the sizes of $f^{3}_{x}$  and $f^{4}_{x}$  are {24, 12}, respectively. The divided window's pixel step size can be set to 4 and 2, respectively, resulting in a significant improvement in computational speed, as can be seen in the second and third rows. Finally, the experimental results of our proposed RFMNet are relatively sensitive to input images within a specific size range. Due to the feature matching via the overlapped windows cross-attention mechanism, regardless of whether the backbone network is ResNet-50 or Swin-S, the model's ability to detect camouflaged objects becomes significantly stronger as the input image size increases. However, it also requires more computational resources correspondingly.

\subsection{Future work}
Given our proposed RFMNet model, there are three significant aspects that warrant in-depth exploration for the progress of Ref-COD research.

{\bf{More comprehensive Ref-COD dataset:}} The proposal of the R2C7K dataset has significantly propelled the advancement of research in Ref-COD. However, in the Camo-subset of this dataset, the majority of images depicting either a single camouflaged object or multiple camouflaged objects belong to a single category. The proportion of images containing two or more categories within a single picture is relatively low. Moreover, it is challenging to obtain images with scenes of two or more categories of camouflaged objects via the Internet. In subsequent research, large models can be employed to generate such images, thereby further facilitating the progress of Ref-COD.

{\bf{Integration of two reference methods:}}
In the Ref-COD task, textual references are characterized by their flexible and diverse semantic descriptions, which enable dynamic representations of camouflaged objects. In contrast, image references provide richer, fine-grained visual cues directly. The effective integration of these complementary modalities has the potential to significantly advance research in both Ref-COD and multimodal learning.

{\bf{Advanced local region matching:}}
Although our proposed overlapped windows cross-attention improves camouflaged object detection, superior regional feature matching methods exist to exploit salient object information more effectively.

\section{Conclusion}
This paper proposes a new feature fusion method for the Ref-COD task. We exploit the richer features of reference information and propose a multi-context feature fusion architecture, which can effectively improve the model's performance in both the reference image and text tasks. Besides, we propose an overlapped windows cross-attention mechanism to make the module pay more attention to local area matching based on reference salient image features. In addition, we propose the referring feature aggregation (RFA) module to progressively aggregate the features layer by layer. Extensive experiment results indicate that our proposed RFMNet can locate the camouflaged object and generate detailed segmentation results. Our proposed ideas will offer inspiration for COD and other related future works. 

\section{Acknowledgement}
This work is supported by National Nature Science Foundation of China (Grant No. 62203291).

\bibliographystyle{elsarticle-num}

\end{document}